\documentclass[11pt,a4paper]{article}
\usepackage[hyperref]{acl2020}
\usepackage{times}
\usepackage{latexsym}
\usepackage{amsmath,amssymb,bbm}
\usepackage{bm}
\usepackage{graphicx}
\usepackage{multirow}
\usepackage{arydshln}
\usepackage{url}

\usepackage{microtype}

\aclfinalcopy 

\title{Probing Contextualized Sentence Representations with Visual Awareness}

\author{Zhuosheng Zhang\textsuperscript{1,2,3}, Rui Wang\textsuperscript{4},  Kehai Chen\textsuperscript{4}, Masao Utiyama\textsuperscript{4}, \\
\large \textbf{Eiichiro Sumita\textsuperscript{4}, Hai Zhao\textsuperscript{1,2,3}}\\
\textsuperscript{1} Department of Computer Science and Engineering, Shanghai Jiao Tong University\\
\textsuperscript{2} Key Laboratory of Shanghai Education Commission for Intelligent Interaction\\
and Cognitive Engineering, Shanghai Jiao Tong University, Shanghai, China\\
\textsuperscript{3}MoE Key Lab of Artificial Intelligence, AI Institute, Shanghai Jiao Tong University, Shanghai, China\\
\textsuperscript{4} National Institute of Information and Communications Technology (NICT)\\
\texttt{zhangzs@sjtu.edu.cn,zhaohai@cs.sjtu.edu.cn}\\
\texttt{\{wangrui,khchen,mutiyama,eiichiro.sumita\}@nict.go.jp}\\
}

\date{}

\begin{document}
\maketitle
\begin{abstract}
  We present a universal framework to model contextualized sentence representations with visual awareness that is motivated to overcome the shortcomings of the multimodal parallel data with manual annotations. For each sentence, we first retrieve a diversity of images from a shared cross-modal embedding space, which is pre-trained on a large-scale of text-image pairs. Then, the texts and images are respectively encoded by transformer encoder and convolutional neural network. The two sequences of representations are further fused by a simple and effective attention layer. The architecture can be easily applied to text-only natural language processing tasks without manually annotating multimodal parallel corpora. We apply the proposed method on three tasks, including neural machine translation, natural language inference and sequence labeling and experimental results verify the effectiveness.
\end{abstract}

\section{Introduction}

Learning vector representations of sentence meaning is a long-standing objective in natural language processing (NLP) \cite{wang2018sentence,chen2019neural,zhang2019sg}. Text representation learning has evolved from word-level distributed representations \cite{mikolov2013distributed,pennington2014glove} to contextualized language modeling (LM) \cite{Peters2018ELMO,radford2018improving,devlin2018bert,yang2019xlnet}. Despite the success of LMs, NLP models are impoverished compared to humans due to the monotonous learning solely from textual features without grounding in the outside world such as visual conception. Therefore, there emerges a trend of researches that are motivated to apply 
non-linguistic modalities into language representations \cite{bruni2014multimodal,calixto2017doubly,zhang2018image,ive2019distilling,shi2019visually}. 

Previous work mainly integrates the visual guidance to word or character representations \cite{kiela2014learning,silberer2014learning,zablocki2018learning,wu2019glyce}, which requires the alignment of word and images. Besides, word meaning may vary in different sentences depending on the context, thus the aligned image would not be optimal. Recently, there is a recent trend of pre-training visual-linguistic (VL) representations for visual and language tasks \cite{su2019vl,lu2019vilbert,tan2019lxmert,li2019unicoder,zhou2019unified,sun2019videobert}. However, these VL studies rely on the text-image annotations as the paired input, thus are retrained only in VL tasks, such as image caption and visual questions answering. However, for NLP tasks, most texts are unlabeled. Therefore, it is essential to probe a general method to apply visual information to a wider range of mono-modal text-only tasks.

Recent studies have verified that the representations of images and texts can be jointly leveraged to build visual-semantic embeddings  in a shared representation space \cite{frome2013devise,karpathy2015deep,ren2016joint,mukherjee2016gaussian}. To this end, a popular approach is to connect both of the mono-modal text and image encoding paths with fully connected layers \cite{wang2018learning,engilberge2018finding}. The shared deep embedding can be used for cross-modal retrieval thus it can associate sentence texts with associated images. Inspired by this line of research, we are motivated to incorporate the visual awareness into sentence modeling by retrieving a group of images for a given sentence.

According to the Distributional Hypothesis \cite{harris1954distributional} which states that words that occur in similar contexts tend to have similar meanings, we make the attempt to extend it to visual modalities, \emph{the sentences with similar meanings would be likely to pair with similar or the same images in the shared embedding space}. In this paper, we propose an approach to model contextualized sentence representations with visual awareness. For each sentence, we retrieve a diversity of images from a shared text-visual embedding space that is pre-trained on a large-scale of text-image pairs to connect both the mono-modal paths of text and image embeddings. The texts and images are encoded by transformer LM and pre-trained convolutional neural network (CNN), respectively. A simple and effective attention layer is then designed to fuse the two sequences of representations. In particular, the proposed approach can be easily applied to text-only tasks without manually annotating multimodal parallel corpora. The proposed method was evaluated on three tasks, including neural machine translation (NMT), natural language inference (NLI) and sequence labeling (SL). Experiments and analysis show effectiveness. In summary, our contributions are primarily three-fold:
\begin{enumerate}
    \item We present a universal visual representation method that overcomes the shortcomings of the multimodal parallel data with manual annotations.
    \item We propose a multimodal context-driven model to jointly learn sentence-level representations from textual and visual modalities.
    \item Experiments on different tasks verified the effectiveness and generality of the proposed approach.
\end{enumerate}

\begin{figure}
	\centering
	\includegraphics[width=0.5\textwidth]{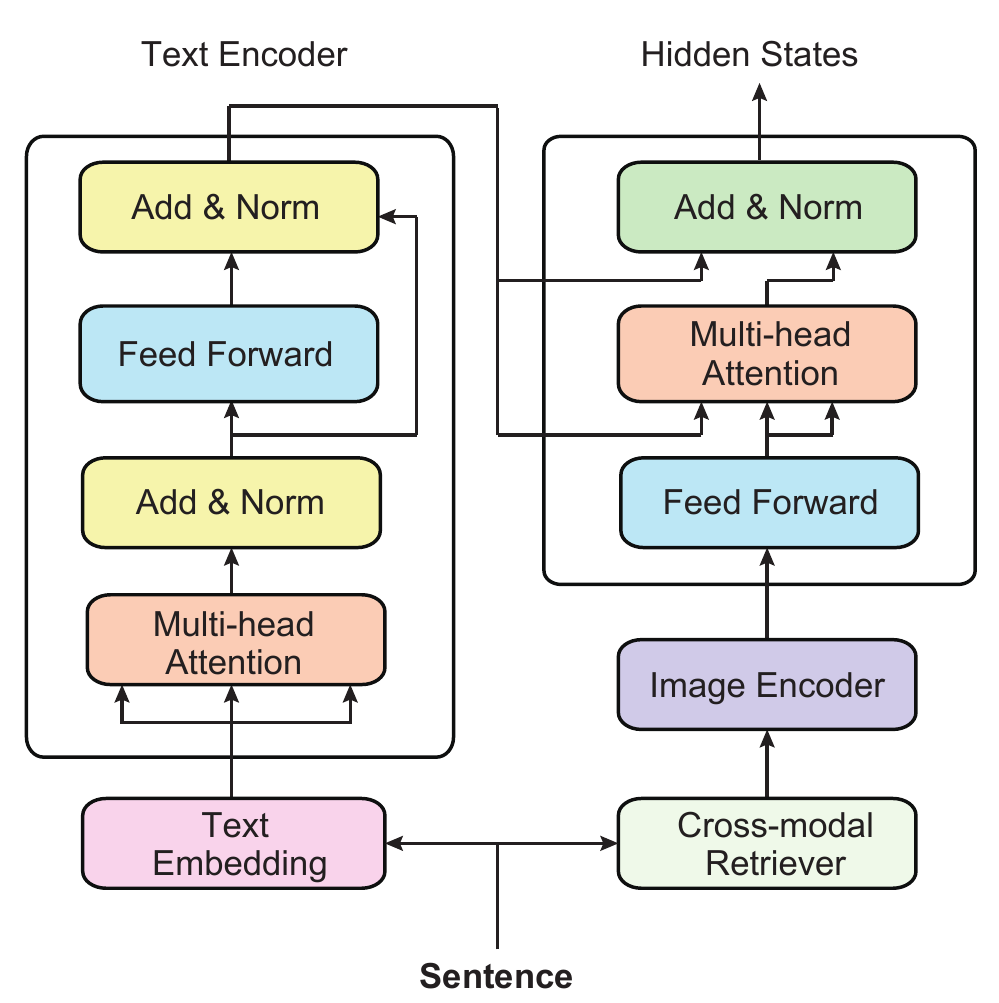}
	\caption{Overview of the our model architecture.}
	\label{overview}
\end{figure}

\section{Approach}
Figure \ref{overview} illustrates the architecture of our proposed method. Given a sentence, we first fetch a group of matched images from the cross-modal retrieval model. The text and images are encoded by the text feature extractor and image feature extractor, respectively. Then the two sequences of representation are integrated by multi-head attention to form a joint representation which is passed to downstream task-specific layers. Before introducing our visual-aware model, let us briefly show the cross-modal retrieval model which is used to for image retrieval given sentence text.

\subsection{Cross-modal Retrieval Model}
\begin{figure*}
	\centering
	\includegraphics[width=1\textwidth]{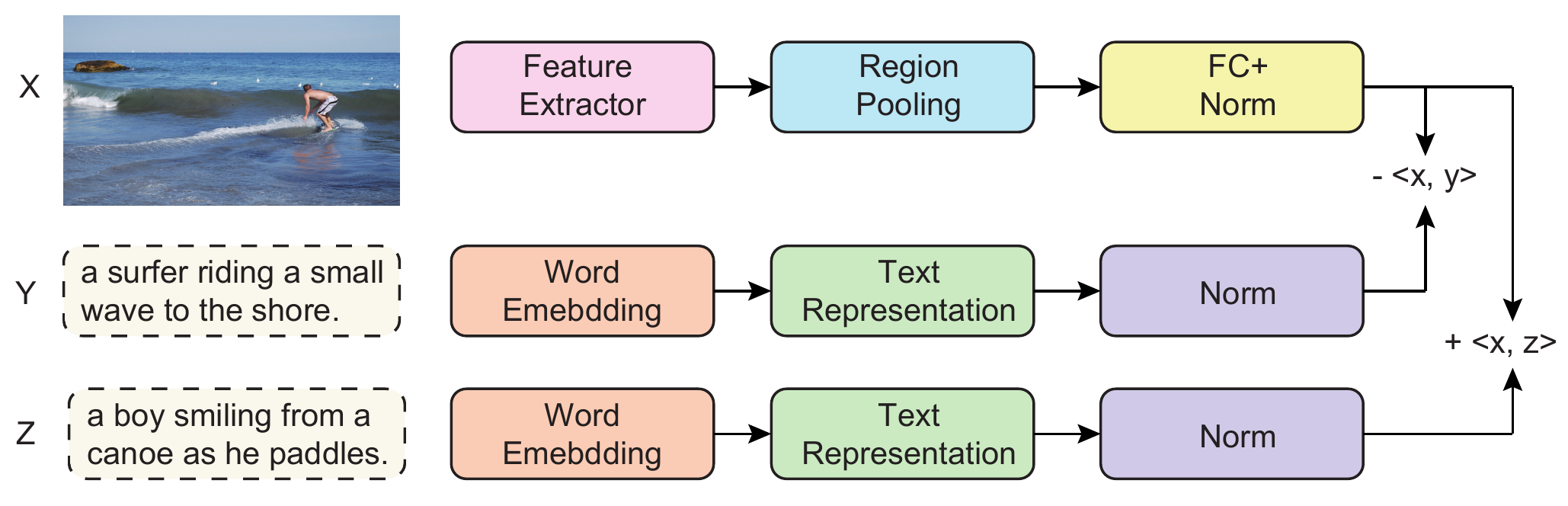}
	\caption{Details of the proposed semantic-visual embedding model.}\label{mapping}
\end{figure*}

For input sentence, our aim is to associate it with a number of images from a candidate corpus. Following \cite{engilberge2018finding}, we train a semantic-visual embedding on a text-image corpus, which is then used for image retrieval. The semantic-visual embedding architecture comprises two paths to encoder the texts and images into vectors, respectively. Based on our preliminary experiments, we choose the simple recurrent unit (SRU) architecture as our text encoder and the fully convolutional residual ResNet-152 \cite{xie2017aggregated} with Weldon pooling  \cite{durand2016weldon}. 

Both pipelines are learned simultaneously, each image is paired with 1) a positive text $Y$ that describes the image $X$ and 2) a hard negative $Z$ which is selected as the one that has the highest similarity with the image while not being associated with it. The architecture of the model is shown on Figure \ref{mapping}. During training, a triplet loss \cite{wang2014learning,schroff2015facenet,gordo2017end} is used to converge correctly and increase our performances as shown in equation~\ref{eq:cost}. 
\begin{equation}
    \label{eq:cost}
    loss(x,y,z) = max(0, \alpha - x \cdot y + x \cdot z),
\end{equation}
where $x$, $y$, and $z$ are respectively the embeddings of $X$, $Y$, and $Z$. $\alpha$ is the minimum margin between the similarity of the correct caption and the unrelated caption. The loss function enables that the sentence $Y$ should be closer to the corresponding image $X$ than the unrelated one $Z$. 

In prediction time, the relationship of texts and images is calculated by cosine similarity. 

\subsection{Visual-aware Model}
\subsection{Encoding Layer}
For each sentence $X=\{x_1, x_2, \dots, x_I\}$, we pair it with the top matched $m$ images $E = \{e_1, e_2, \dots, e_m\}$ according to the retrieval method above.
Then, the sentence $X$=$\{x_1, x_2, \dots, x_I\}$ is fed into multi-layer transformer encoder to learn the sentence representation $\textbf{H}$.
Meanwhile, the images $E$ =$\{e_1, e_2, \dots, e_m\}$ are encoded by a pre-trained ResNet \citep{he2016deep} followed by a feed forward layer to learn the image representation $\textbf{M}$ with the same dimension with $\textbf{H}$.

\subsection{Multi-modal Integration Layer}
Then, we apply a one-layer multi-head attention mechanism to append the image representation to the text representation:
\begin{equation}
    \centering
	 \bm{\textbf{H}}' = \textup{ATT}(\textbf{H}, \textbf{K}_{\textbf{M}}, \textbf{V}_{\textbf{M}}), 
\label{eq2:Image_Representation}
\end{equation}
where \{$\textbf{K}_{\textbf{M}}$, $\textbf{V}_{\textbf{M}}$\} are packed from the learned image representation $\bm{\textbf{M}}$.

Following the same practice in the transformer block, We fuse $\textbf{H}$ and $\overline{\bm{\textbf{H}}}$ with layer normalization to learn the joint representation:
\begin{equation} \bm{\hat{\textbf{H}}}=\textup{LayerNorm}(W(\textbf{H}+\bm{\textbf{H}}')+b).
\label{eq4:Fusing_Sourece_Representation}
\end{equation}
where $W$ and $b$ are parameters.
\subsection{Task-specific Layer}
In this section, we show how the joint representation is used for downstream tasks by taking NMT, NLI and SL tasks for example. For NMT, $\hat{\textbf{H}}$ is fed to the decoder to learn a dependent-time context vector for predicting target translation. For NLI and SL, $\hat{\textbf{H}}$ is directly fed to a feed forward layer to make the prediction.

\section{Task Settings}
We evaluate our model on three different NLP tasks, namely neural machine translation, natural language inference and sequence labeling. We present these evaluation benchmarks in what follows.
\subsection{Neural Machine Translation}
We use two translation datasets, including WMT'16 English-to-Romanian (EN-RO), and WMT'14 English-to-German (EN-DE) which are standard corpora for NMT evaluation.

1) For the EN-RO task, we experimented with the officially provided parallel corpus: Europarl v7 and SETIMES2 from WMT'16 with 0.6M sentence pairs. We used \textit{newsdev2016} as the dev set and \textit{newstest2016} as the test set.

2) For the EN-DE translation task, 4.43M bilingual sentence pairs of the WMT14 dataset were used as training data, including Common Crawl, News Commentary, and Europarl v7. 
The \textit{newstest2013} and \textit{newstest2014} datasets were used as the dev set and test set, respectively.

\subsection{Natural Language Inference}
Natural Language Inference involves reading a pair of sentences and judging the relationship between their meanings, such as entailment, neutral and contradiction. In this task, we use the  Stanford  Nat-ural  Language  Inference  (SNLI)  corpus  (Bowmanet al., 2015) which provides approximately 570k hypothesis/premise pairs. 

\subsection{Sequence Labeling}
We use the CoNLL-2003 Named Entity Recognition dataset \cite{tjong2003introduction} for the sequence labeling task, which includes four kinds of NEs: Person, Location, Organization and MISC.

\section{Model Implementation}\label{sec:imp}
Now, we introduce the specific implementation parts of our method. All the experiments were done on 8 NVIDIA TESLA V100 GPUs.

\subsection{Cross-modal Retrieval Model}\label{srl}
The cross-modal retrieval model is trained on the MS-COCO dataset \cite{lin2014microsoft} that contains 123 287 images with 5 English captions per image. It is split into
82,783 training images, 5,000 validation images and 5,000
testing images. We used the Karpathy split \cite{karpathy2015deep} that forms 113, 287 training, 5,000 validation and 5,000 test images. The model is implemented following the same settings in \cite{engilberge2018finding} with the state-of-the-art results (94.0\% R@10) in cross-modal retrieval. The maximum number of retrieved images $m$ for each sentence is set to 8 according to our preliminary experimental results. 

\begin{table}
	\resizebox{\linewidth}{!}{
	\begin{tabular}{lll}
		\hline
		
		\hline
		\textbf{Model} &  \textbf{EN-RO} & \textbf{EN-DE}   \\ 
		\hline
		\multicolumn{3}{c}{\emph{Public Systems}} \\
		Trans. \cite{NIPS2017_7181} & -  & 27.3 \\
		Trans. \cite{lee-etal-2018-deterministic} & 32.40 & -\\
		\hline
		\multicolumn{3}{c}{\emph{Our implementation}} \\
		Trans. & 32.66  &  27.31 \\
		\textbf{+ VA} & \textbf{34.63} &  \textbf{27.83} \\
		\hline
		
		\hline
	\end{tabular}
	}
	{
		\caption{BLEU scores on EN-RO and EN-DE for the NMT tasks. Trans. is short for transformer.}
\label{tbl:nmt} } 
	
\end{table}

\subsection{Baseline}
To incorporate our visual-aware model (+VA), we only modify the encoder of the baselines  by introducing the image encoder layer and multi-modal integration layer.

For the NMT tasks, the baseline was Transformer \citep{NIPS2017_7181} implemented by \emph{fairseq}\footnote{\url{https://github.com/pytorch/fairseq}} \citep{ott2019fairseq}. We used six layers for the encoder and the decoder. The number of dimensions of all input and output layers was set to 512. The inner feed-forward neural network layer was set to 2048. The heads of all multi-head modules were set to eight in both encoder and decoder layers. The byte pair encoding algorithm was adopted, and the size of the vocabulary was set to 40,000.
In each training batch, a set of sentence pairs contained approximately 4096$\times$4 source tokens and 4096$\times$4 target tokens. 
During training, the value of label smoothing was set to 0.1, and the attention dropout and residual dropout were \textit{p} = 0.1. 
The Adam optimizer~\citep{DBLP:journals/corr/KingmaB14} was used to tune the parameters of the model. The learning rate was varied under a warm-up strategy with 8,000 steps.

For the NLI and SL tasks, the baseline was BERT (Base)\footnote{\url{https://github.com/huggingface/pytorch-pretrained-BERT}}. We used the pre-trained weights of BERT and follow the same fine-tuning procedure as BERT without any modification. The initial learning rate was set in \{8e-6, 1e-5, 2e-5, 3e-5\} with warm-up rate of 0.1 and L2 weight decay of 0.01. The batch size is selected in \{16, 24, 32\}. The maximum number of epochs is set in [2, 5]. Texts are tokenized using wordpieces, with maximum length of 128.

\subsection{Results}

Table~\ref{tbl:nmt} shows the translation results for the WMT'14 EN-DE and WMT'16 EN-RO translation task.
We see that our method significantly outperformed the baseline Transformer, demonstrating the effectiveness of modeling visual information for NMT.

\begin{table}
	\centering
	\begin{tabular}{p{5.8cm}p{1cm}}
		\hline
		
		\hline
		\textbf{Model} &  \textbf{Acc}   \\ 
		\hline
		\multicolumn{2}{c}{\emph{Public Systems}} \\
		GPT \cite{radford2018improving}  & 89.9 \\
		DRCN \cite{kim2018semantic} & 90.1\\
		MT-DNN \cite{liu2019multi} & 91.6\\
		SemBERT \cite{zhang2019semantics} & 91.6 \\
		\hdashline
		BERT (Base) \cite{liu2019multi} & 90.8 \\
		\hline
		\multicolumn{2}{c}{\emph{Our implementation}} \\
		BERT (Base) & 90.7\\
		\textbf{+ VA} & \textbf{91.2} \\
		\hline
		
		\hline
	\end{tabular}
	{
		\caption{\label{tab:snli} Accuracy on SNLI dataset.} } 
	
\end{table}

	\begin{table}
	\resizebox{\linewidth}{!}
		{
			
			\begin{tabular}{lccc}
				\hline
				
				\hline
				
				\textbf{Model}& \textbf{F1 score} \\
				\multicolumn{2}{c}{\emph{Public Systems}} \\
				LSTM-CRF \cite{lample2016neural} & 90.94 \\
				\hdashline
				BERT (Base) \cite{pires2019multilingual} & 91.07 \\
				\hline
				\multicolumn{2}{c}{\emph{Our implementation}} \\
				BERT (Base) &  91.21 \\
				\textbf{+VA} &   \textbf{91.46} \\
				\hline
				
				\hline
			\end{tabular}
		}
		\caption{Results (\%) of CoNLL-2003 NER dataset.}
		\label{table-ner-result}
		\centering
	\end{table}

\begin{figure*}
	\centering
	\includegraphics[width=1.0\textwidth]{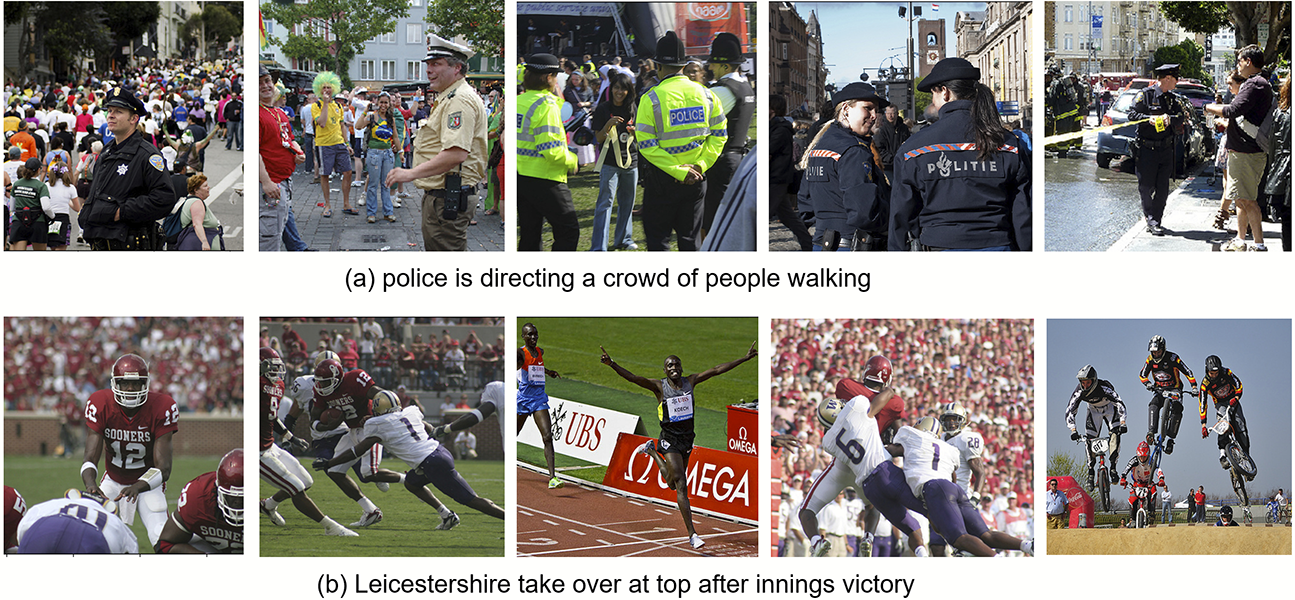}
	\caption{Examples of the retrieved images for sentences.}
	\label{exp_re}
\end{figure*}

Table~\ref{tab:snli}-\ref{table-ner-result} show the results for the NLI and SL tasks, which also verify the effectiveness. The results show the our method is not only useful for the fundamental tagging task but also more advanced translation and inference tasks.

\section{Analysis}

\subsection{Concept Localization}
\begin{figure}
	\centering
	\includegraphics[width=0.49\textwidth]{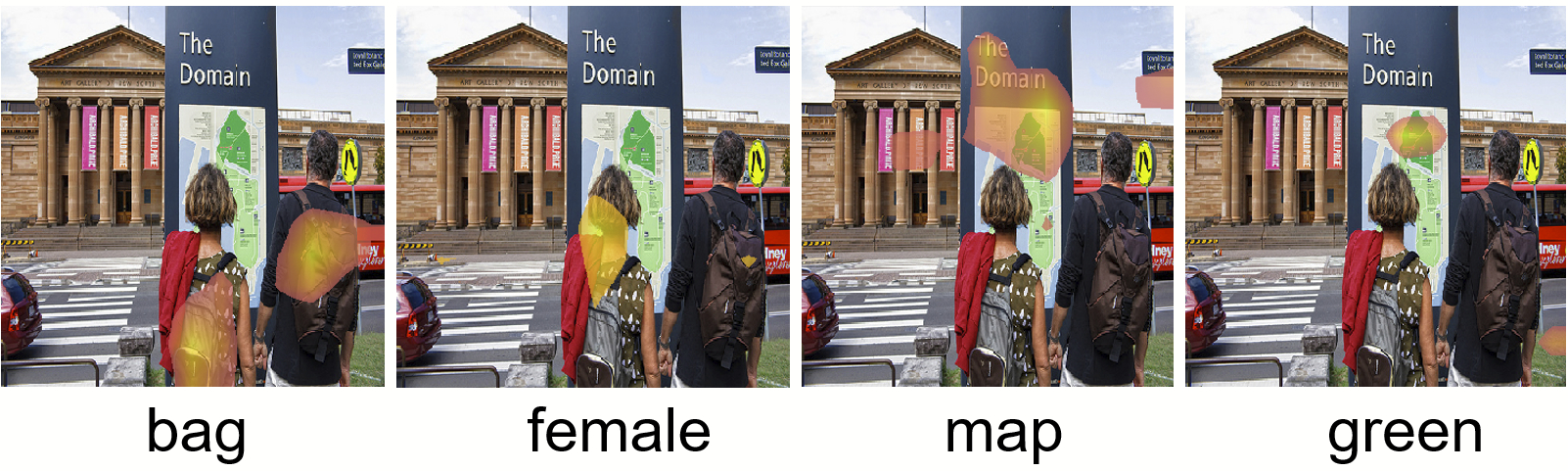}
	\caption{Concept activation maps with different input words. The orange region indicates the highest peak in the heatmap. 
}
	\label{local}
\end{figure}

We observe an important advantage of shared embedding space is that it can address the localization of arbitrary concepts within the image. For input text, we compute the image localization heatmap derived from the activation map of the last convolutional layer following \cite{engilberge2018finding}. Figure \ref{local} shows the example, which indicates that the shared space can not only perform the image retrieval, but also match language concepts in the image for any text query.

\subsection{Examples of Image Retrieval}
Although the image retrieval method achieves wonderful results on COCO datasets, we are interested in the explicit results on our task-specific datasets. We randomly select some examples to interpret the image retrieval process intuitively, as shown in Figure \ref{exp_re}. Besides the ``good" examples that show good matched contents of the text and image, we also observe some ``negative" examples that the contents might not be related in concept but show some potential connections. To some extent, the alignment of the text and image concepts might not be the only effective factor for multimodal modeling, since they are defined by human knowledge and the meanings may vary among different people or different time. In contrast, the consistent mapping relationships of the modalities in a shared embedding space would be more potentially beneficial, because the similar images tend to be retrieved for similar sentences, which can play the role of topical hints for sentence modeling.

\section{Conclusion}
In this work, we present a universal method to incorporate visual information into sentence modeling by conducting image retrieval from a pre-trained shared cross-modal embedding space to overcome the shortcomings of the manual annotated multimodal parallel data. The text and image representations are respectively encoded by transformer encoder and convolutional neural network and then integrated in a multi-head attention layer. Empirical studies on a wide range of NLP tasks including NMT, NLT and SL verify the effectiveness. Our method is general and fundamental and can be easily implemented to any existing deep learning NLP system. We hope this work will facilitate future multimodal researches across vision and language.

\bibliography{acl2020}
\bibliographystyle{acl_natbib}

\end{document}